\def\year{2022}\relax
\pdfoutput=1
\documentclass[letterpaper]{article} 
\usepackage{aaai22}  
\usepackage{times}  
\usepackage{helvet}  
\usepackage{courier}  
\usepackage[hyphens]{url}  
\usepackage{graphicx} 
\usepackage{natbib}  
\usepackage{caption} 
\usepackage{multirow}
\usepackage{booktabs}
\usepackage{color}
\usepackage{amssymb}
\usepackage{bbding}
\usepackage{amsmath}
\usepackage{diagbox}
\DeclareCaptionStyle{ruled}{labelfont=normalfont,labelsep=colon,strut=off} 
\usepackage{algorithm}
\usepackage{algorithmic}
\usepackage{url}
\usepackage[pagebackref=true,breaklinks=true,letterpaper=true,colorlinks,bookmarks=false]{hyperref}

\usepackage{newfloat}
\usepackage{listings}
\floatstyle{ruled}
\newfloat{listing}{tb}{lst}{}
\floatname{listing}{Listing}
\title{Efficient Non-Local Contrastive Attention for Image Super-Resolution}
\author {
    Bin Xia\textsuperscript{\rm 1}\equalcontrib, 
    Yucheng Hang\textsuperscript{\rm 1}\equalcontrib, 
    Yapeng Tian\textsuperscript{\rm 2}, 
    Wenming Yang\textsuperscript{\rm 1}\thanks{Corresponding  author.}, Qingmin Liao\textsuperscript{\rm 1}, Jie Zhou\textsuperscript{\rm 1}
}
\affiliations{
    \textsuperscript{\rm 1} Tsinghua University\\
    \textsuperscript{\rm 2} University of Rochester\\

    xiab20@mails.tsinghua.edu.cn, hangyc20@mails.tsinghua.edu.cn, yapengtian@rochester.edu, yang.wenming@sz.tsinghua.edu.cn, liaoqm@tsinghua.edu.cn,  jzhou@tsinghua.edu.cn
}

\usepackage{bibentry}

\begin{document}

\maketitle

\begin{abstract}
Non-Local Attention (NLA) brings significant improvement for Single Image Super-Resolution (SISR) by leveraging intrinsic feature correlation in natural images. However, NLA gives noisy information large weights and consumes quadratic computation resources with respect to the input size, limiting its performance and application. In this paper, we propose a novel Efficient Non-Local Contrastive Attention (ENLCA) to perform long-range visual modeling and leverage more relevant non-local features. Specifically, ENLCA consists of two parts, Efficient Non-Local Attention (ENLA) and Sparse Aggregation. ENLA adopts the kernel method to approximate exponential function and obtains linear computation complexity. For Sparse Aggregation, we multiply inputs by an amplification factor to focus on informative features, yet the variance of approximation increases exponentially. Therefore, contrastive learning is applied to further separate relevant and irrelevant features. To demonstrate the effectiveness of ENLCA, we build an architecture called Efficient Non-Local Contrastive Network (ENLCN) by adding a few of our modules in a simple backbone. Extensive experimental results show that ENLCN reaches superior performance over state-of-the-art approaches on both quantitative and qualitative evaluations. The code is available at \url{https://github.com/Zj-BinXia/ENLCA}. 
\end{abstract}

\section{Introduction}

Single image super-resolution (SISR) has essential applications in certain areas, such as surveillance monitoring and medical detection. The goal of SISR is to generate a high-resolution (HR) image with realistic textures from its low-resolution (LR) counterpart. However, SISR is an ill-posed inverse problem, which is challenging to produce high-quality HR details. Thus, numerous image priors \cite{{prior1},{prior2},{prior3},{glasner2009super}} are introduced to limit the solution space of SR, including local and non-local prior.

Among traditional methods, non-local priors \cite{{glasner2009super},{zontak2011internal}} have been widely used. NCSR \cite{self_sim} reconstructs SR pixels by the weighted sum of similar patches in the LR image itself. Besides, \citet{huang2015single} expands the internal similar patch search space by allowing geometric variations. 

\begin{figure}[tb]
	\centering
	\includegraphics[height=2.4cm,width=8cm]{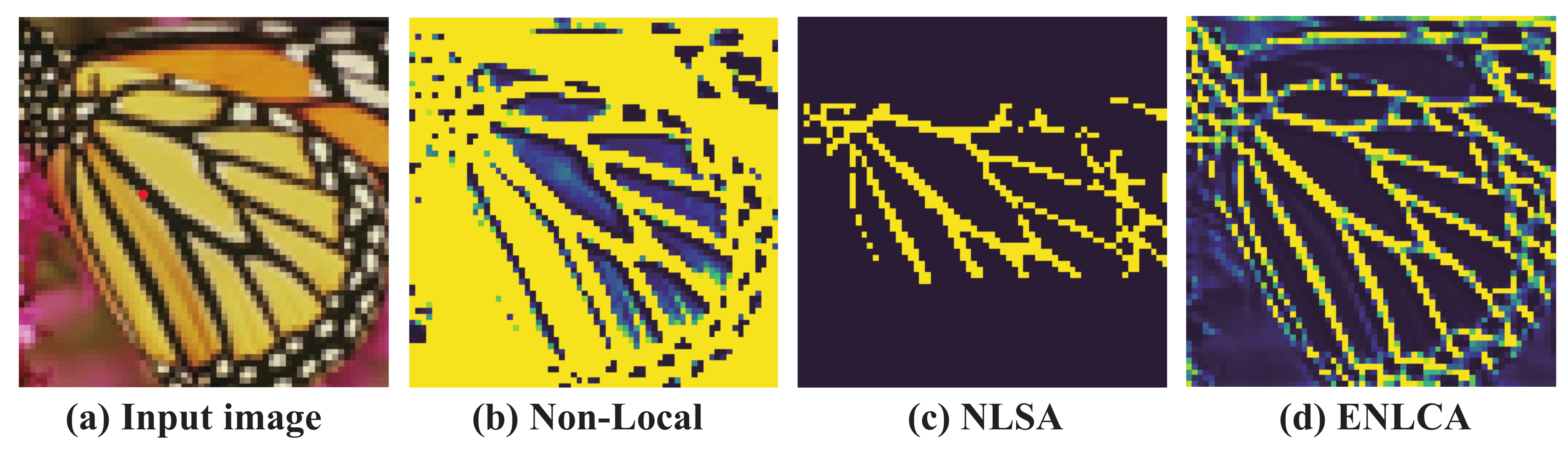} 
	\caption{The visualization of correlation maps obtained by softmax and inner product between the marked red feature and all features. We can see that Non-Local attention pays attention to irrelevant features, and NLSA \cite{NLSN} ignores the related features, while our ENLCA can simultaneously leverage relevant features and suppress irrelevant contents for SR.}
	\label{fig:heatmap}
\end{figure}
Since SRCNN \cite{SRCNN} firstly adopted deep learning in SISR, deep learning based methods have achieved a significant performance boost compared with traditional methods. The key to the success of deep learning is learnable feature representation. Therefore, certain works increase the depth and width of the network to further expand the receptive field and enhance representation ability. However, the solutions merely utilize relevant local information. Afterward, networks equip non-local modules, formulated as Eq \ref{equal:non-local}, to exploit the image self-similarity prior globally and boost performance. Nevertheless, as shown in Figure \ref{fig:heatmap} (b), the non-local module fuses excessive irrelevant features, which imports noise and limits the non-local module performance. In addition, the non-local module requires to compute feature mutual-similarity among all the pixel locations, which leads to quadratic computational cost to the input size. Limiting the non-local matching range can alleviate the issue at the expense of the loss of much global information. To address the problems, NLSA \cite{NLSN} adopted Locality Sensitive Hashing (LSH) to aggregate the possible relevant features rapidly. However, as shown in Figure \ref{fig:heatmap} (c), NLSA fixes the maximum number of hash buckets and chunk sizes to keep linear complexity with the input size, which results in ignoring important global related information.

In this paper, we aim to aggregate all important relevant features, keep the sparsity in the non-local module, and largely reduce its computational cost. Consequently, we propose a novel Efficient Non-Local Contrastive Attention (ENLCA) module and embed it into the deep SR network like EDSR. Specifically, we propose an Efficient Non-Local Attention (ENLA) module with kernel function approximation and associative law of matrix multiplication. The ENLA module achieves comparable performance as the standard non-local module while merely requiring linear computation and space complexity with respect to the input size. To further improve the performance of ENLA, we give the related features larger weights and ignore unrelated features in the aggregation process by multiplying the query and key by an amplification factor $k$. However, the kernel function of ENLA is based on Gaussian random vector approximation, and the variance of approximation increases exponentially with the inner product of query and key amplifying. Thus, we apply contrastive learning on ENLA to further increase the distance between relevant and irrelevant features. The contributions of our paper can be summarized as follows:  
   
\begin{itemize}
	\item We propose a novel Efficient Non-Local Contrastive Attention (ENLCA) for the SISR task. The ENLA of ENLCA significantly reduces the computational complexity from quadratic to linear by kernel function approximation and associative law of matrix multiplication.
	\item We enforce the aggregated feature sparsity by amplifying the query and key. In addition, we apply contrastive learning on the ENLA to further strengthen the effect of relevant features. 
	\item A few ENLCA modules can improve a fairly simple ResNet backbone to state-of-the-arts. Extensive experiments demonstrate the advantages of ENLCA over the standard Non-Local Attention (NLA) and Non-Local Sparse Attention.
\end{itemize}

\section{Related Work}

\subsection{Non-Local Attention in Super-Resolution}
The SISR methods\cite{{SRCNN},{VDSR},{SRGAN},{ESRGAN}} based on deep learning learn an end-to-end image mapping function between LR and HR images and obtain superior performance to conventional algorithms. In recent years, to further improve the performance of models, there is an emerging trend of applying non-local attention. Methods, such as CSNLN \cite{CSNLN}, SAN \cite{SAN}, RNAN \cite{RNAN}, NLRN \cite{NLRN}, inserted non-local attention into their networks to make full use of recurring small patches and achieved considerable performance gain. However, the existing NLAs developed for SISR problems consume excessive computational resources and fuse much noisy information. Hence, NLSA \cite{NLSN} uses Locality Sensitive Hashing (LSH) to efficiently aggregate several relevant features. Nevertheless, NLSA may miss the informative features, and computational cost can be further optimized.  Motivated by recent work \cite{{reformer},{random_feature},{performer}} on self-attention methods for natural language processing, we propose Efficient Non-Local Contrastive Attention (ENLCA) to learn global feature relations and reduce computational complexity.  

\subsection{Contrastive Learning}
Contrastive learning has been widely studied for unsupervised representation learning in recent years. Instead of minimizing the difference between the output and a fixed target, contrastive learning \cite{{chen2020simple},{he2020momentum},{henaff2020data},{oord2018representation}} maximizes the mutual information in representation space. Nevertheless, different from high-level vision tasks \cite{{wu2018unsupervised},{he2020momentum}}, there are few works to apply contrastive learning on low-level vision tasks. Recently, contrastive learning has been adopted in BlindSR\cite{{CRL-SR},{DASR}} to distinguish different degradations and achieved significant improvements. To the best of our knowledge, we are the first to introduce contrastive learning to the non-local module for enhancing sparsity by pulling relevant features close and pushing irrelevant away in representation space.                                     

\section{Efficient Non-Local Contrastive Attention}
\begin{figure*}[htb]
	\centering
	\includegraphics[height=6.7cm,width=18cm]{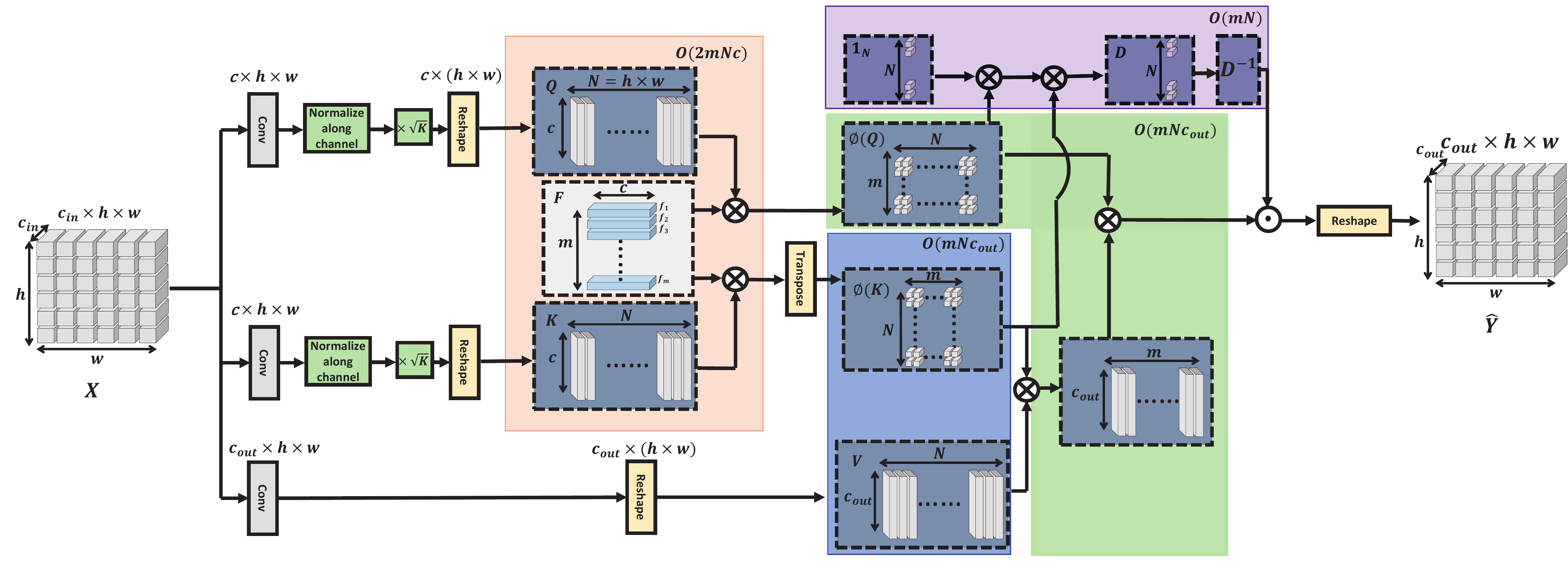} 
	\caption{The illustration of Efficient Non-Local Attention. $h$ and $w$ are the input image height and width. $c_{in}$, $c$, and $c_{out}$ are the number of channels. $m$ indicates the number of random samples, and $N$ is the input size. $\boldsymbol{Q}$ and $\boldsymbol{K}$ are feature maps extracted from input $\boldsymbol{X}$.  $\boldsymbol{F}$ is an Gaussian random matrix to transform $\boldsymbol{Q}$ and $\boldsymbol{K}$ to $\phi(\boldsymbol{Q})$ and $\phi(\boldsymbol{K})$, respectively.  $\phi(\boldsymbol{Q})^{T}\phi(\boldsymbol{K})$ is the approximation of $exp(\boldsymbol{Q}^{T}\boldsymbol{K})$. $\boldsymbol{V}$ multiply $\phi(\boldsymbol{K})$ and $\phi(\boldsymbol{Q})$ successively to generate the features aggregating global information with linear computational complexity to the input size.} 
	\label{fig:kernel_nonlocal}
\end{figure*}

This section introduces the proposed Efficient Non-Local Contrastive Attention (ENLCA). The module realizes important global relevant information aggregation at the cost of linear complexity to the input size. Firstly, we develop Efficient Non-Local Attention (ENLA), the architecture of which is shown in Figure \ref{fig:kernel_nonlocal}. Subsequently, as shown in Figure \ref{fig:contrastive_method}, we introduce Sparse Aggregation by multiplying the inputs an amplification factor and using contrastive learning for ENLCA to further filter noisy information. Finally, we utilize EDSR as the backbone to demonstrate the effectiveness of our module.

\subsection{Efficient Non-Local Attention}
Standard Non-Local Attention aggregates all features, which could propagate irrelevant noises into restored images. NLSA selects possible relevant features for aggregation by Locality Sensitive Hashing (LSH). However, LSH may ignore useful non-local information since it merely leverages relevant information roughly within limited window sizes.   
To alleviate the issue, we propose Efficient Non-Local Attention to aggregate all features efficiently.

\textbf{Non-Local Attention}.
Non-local attention can explore self-exemplars by aggregating relevant features from the whole image. Formally, non-local attention is defined as:
\begin{equation}
\boldsymbol{Y}_{i}=\sum_{j=1}^{N} \frac{\exp\left(\boldsymbol{Q}_{i}^{T} \boldsymbol{K}_{j}\right)}{\sum_{\hat{j}=1}^{N} \exp\left(\boldsymbol{Q}_{i}^{T} \boldsymbol{K}_{\hat{j}}\right)} \boldsymbol{V}_{j},
\label{equal:non-local}
\end{equation}
\begin{equation}
\boldsymbol{Q}=\theta\left(\boldsymbol{X}\right),\boldsymbol{K}=\delta\left(\boldsymbol{X}\right),\boldsymbol{V}=\psi\left(\boldsymbol{X}\right),
\end{equation}
where  $\boldsymbol{Q}_{i}, \boldsymbol{K}_{j} \in \mathbb{R}^{c}$ and $\boldsymbol{V}_{j} \in \mathbb{R}^{c_{out}}$  are pixel-wise features at location $i$ or $j$ on the feature map $\boldsymbol{Q}$, $\boldsymbol{K}$and $\boldsymbol{V}$ respectively. $ \boldsymbol{Y}_{i} \in \mathbb{R}^{c_{out}}$ is the output at location $i$, $\boldsymbol{X}$ is the input and $N$ is the input size. $\theta(.)$, $\delta(.)$, and $\psi(.)$ are feature transformation functions for the input $\boldsymbol{X}$.

\textbf{Efficient Non-Local Attention}. The architecture of ENLA is shown in Figure \ref{fig:kernel_nonlocal}. We decompose $\exp\left(\boldsymbol{Q}_{i}^{T} \boldsymbol{K}_{j}\right)$ by Gaussian random feature approximation and change multiplication order to obtain linear complexity with respect to image size. The decomposition of the exponential kernel function is derived as follows, and the detailed proofs are given in the supplementary.
\begin{equation}
\boldsymbol{Q}=\sqrt{k}\frac{\theta\left(\boldsymbol{X}\right)}{\|\theta\left(\boldsymbol{X}\right)\|},\boldsymbol{K}=\sqrt{k}\frac{\delta\left(\boldsymbol{X}\right)}{\|\theta\left(\boldsymbol{X}\right)\|},\boldsymbol{V}=\psi\left(\boldsymbol{X}\right),
\label{Eq:sparse}
\end{equation}
\begin{equation}
\begin{aligned}
&\mathrm{K}(\boldsymbol{Q}_{i}, \boldsymbol{K}_{j})=\exp \left(\boldsymbol{Q}_{i}^{\top} \boldsymbol{K}_{j}\right)\\
&= \exp \left(-\left(\|\boldsymbol{Q}_{i}\|^{2} + \| \boldsymbol{K}_{j} \|^{2}\right)/2\right)  \exp \left( \|\boldsymbol{Q}_{i}+\boldsymbol{K}_{j}\|^{2}/2\right)\\
& = \mathbb{E}_{\boldsymbol{f} \sim \mathcal{N}\left(\boldsymbol{0}_{c}, \boldsymbol{I}_{c}\right)} \exp \left(\boldsymbol{f}^{\top}\left( \boldsymbol{Q}_{i}+\boldsymbol{K}_{j}\right)-\frac{\|\boldsymbol{Q}_{i}\|^{2}+\|\boldsymbol{K}_{j}\|^{2}}{2}\right) \\
&= \phi(\boldsymbol{Q}_{i})^{T}\phi(\boldsymbol{K}_{j}),
\end{aligned}
\end{equation}
where $\boldsymbol{X}$ is the input feature map, and amplification factor $k$ $(k>1)$ is used for enforcing non-local sparsity. $\theta(.)$, $\delta(.)$, and $\psi(.)$ are feature transformation.  $\boldsymbol{Q}_{i}$ and $\boldsymbol{K}_{j} \in \mathbb{R}^{c}$  are pixel-wise features at location $i$ or $j$ on the feature map $\boldsymbol{Q}$ and $\boldsymbol{K}\in \mathbb{R}^{c \times N}$. $\boldsymbol{f} \in \mathbb{R}^{c}$ and $\boldsymbol{f} \sim \mathcal{N}\left(\boldsymbol{0_{c}}, \boldsymbol{I_{c}}\right)$. In practice, we set  $m$ different Gaussian random samples $\boldsymbol{f}_{1}, \ldots, \boldsymbol{f}_{m} \stackrel{\text { iid }}{\sim} \mathcal{N}\left(\boldsymbol{0}_{c}, \boldsymbol{I}_{c}\right)$ and concatenate them as an Gaussian random matrix $\boldsymbol{F}\in \mathbb{R}^{m \times c}$. Consequently, the exponential-kernel admits a Gaussian random feature map unbiased approximation with $\phi(\boldsymbol{Q}_{i})^{T}\phi(\boldsymbol{K}_{j})$, where $\phi(\boldsymbol{u})=\frac{1}{\sqrt{m}} \exp \left(-\|\boldsymbol{u}\|^{2} / 2\right)  \exp \left(\boldsymbol{F} \boldsymbol{u}\right)$ for $\boldsymbol{u}\in\mathbb{R}^{c}$ and $\phi(\boldsymbol{u}) \in \mathbb{R}^{m}$. 


Based on the above deduction, the Efficient Non-Local Attention can be expressed as:  
\begin{equation}
\begin{aligned}
\boldsymbol{\hat{Y}}=\boldsymbol{D}^{-1}\left( \phi(\boldsymbol{Q})^{\top}\left(\phi(\boldsymbol{K}) \boldsymbol{V}^{\top}\right)\right),
\end{aligned}
\end{equation}

\begin{equation}
\begin{aligned}
\boldsymbol{D}=\operatorname{diag}\left[\phi(\boldsymbol{Q})^{\top}\left(\phi(\boldsymbol{K}) \boldsymbol{1}_{N}\right)\right],
\end{aligned}
\end{equation}
where $\boldsymbol{\hat{Y}}$ stands for the approximated standard non-local attention, $\boldsymbol{D}$ is the normalization item in the softmax operator, and brackets indicate the order of computations.

We present here the theory analysis of variance of exponential kernel function approximation. The detailed proofs are given in the supplementary.
\begin{equation}
\begin{aligned}
&\operatorname{Var}\left(\phi(\boldsymbol{Q}_{i})^{T}\phi(\boldsymbol{K}_{j})\right)=\\
&\frac{1}{m} \exp \left(-\left(\|\boldsymbol{Q}_{i}\|^{2}+\|\boldsymbol{K}_{j}\|^{2}\right)\right) \operatorname{Var}\left(\exp \left(\boldsymbol{f}^{\top} (\boldsymbol{Q}_{i} + \boldsymbol{K}_{j})\right)\right) \\
&=\frac{1}{m} \mathrm{K}^{2}(\boldsymbol{Q}_{i}, \boldsymbol{K}_{j})  \left(\exp \left(\|\boldsymbol{Q}_{i} + \boldsymbol{K}_{j}\|^{2}\right)-1\right).
\end{aligned}
\end{equation}

\begin{figure*}[t]
	\centering
	\includegraphics[height=3cm,width=18cm]{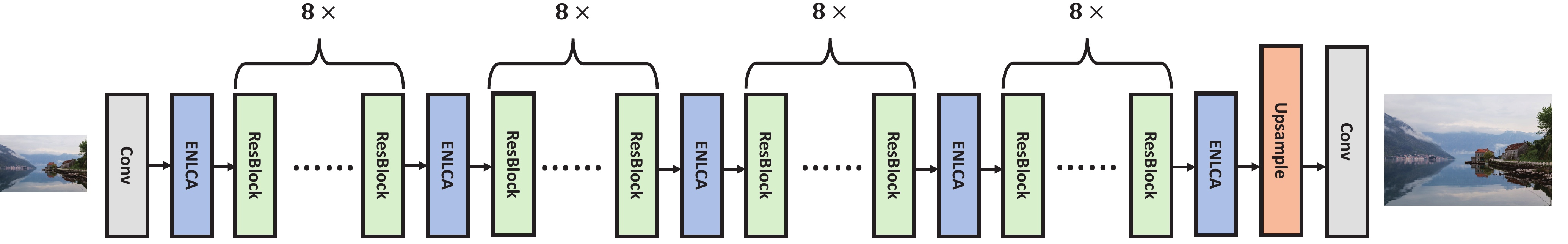} 
	\caption{The proposed Efficient Non-Local Contrastive Network (ENLCN). Five ENLCA modules are embedded after every eight residual blocks.}
	\label{fig:all_process}
\end{figure*}

Consequently, as $\mathrm{K}(\boldsymbol{Q}_{i}, \boldsymbol{K}_{j})$ increases, the variance of $\phi(\boldsymbol{Q}_{i})^{T}\phi(\boldsymbol{K}_{j})$ increases exponentially. To guarantee the accuracy of the approximation results, multiplying $\mathrm{K}(\boldsymbol{Q}_{i}, \boldsymbol{K}_{j})$ by a large amplification factor $k$ is impossible. Additionally, as \cite{orthogonal} did, keep Gaussian random samples orthogonal can reduce the approximation variance. 

\textbf{Computational Complexity.} We analyze the computational complexity of the proposed ENLCA. As shown in Figure \ref{fig:kernel_nonlocal}, the projections from $\boldsymbol{Q}$ and $\boldsymbol{K}$ to $\phi(\boldsymbol{Q})$ and $\phi(\boldsymbol{K})$ by matrix multiplication with $\boldsymbol{F}$ consume $\mathcal{O}(2mNc)$. The cost for the multiplication between $\phi(\boldsymbol{K})$ and $\boldsymbol{V}$ is $\mathcal{O}(mNc_{out})$. Similarly, the cost of the multiplication between $\phi(\boldsymbol{Q})$ and $\phi(\boldsymbol{K}) \boldsymbol{V}^{\top}$ is $\mathcal{O}(mNc_{out})$ as well. Besides, the normalization item  $\boldsymbol{D}$ adds additional $\mathcal{O}(mN)$. Therefore, the overall computational cost of our ENLCA
is $\mathcal{O}(2mNc+2mNc_{out}+mN)$, which only takes linear computational complexity with respect to the input spatial size.

\subsection{Sparse Aggregation}
To further improve the performance of the Efficient Non-Local Attention, we filter out irrelevant information and enlarge the weight of related information. 

Intuitively, multiplying the input by an amplification factor $k (k>1)$ enforces the non-local attention to give higher aggregation weight on related information, the essence of which is to enhance the sparsity of non-local attention weights. Unfortunately, multiplying an amplification factor $k (k>1)$ results in the increment of ENLA approximation variance.

\begin{figure}[htb]
	\centering
	\includegraphics[height=5.6cm,width=8cm]{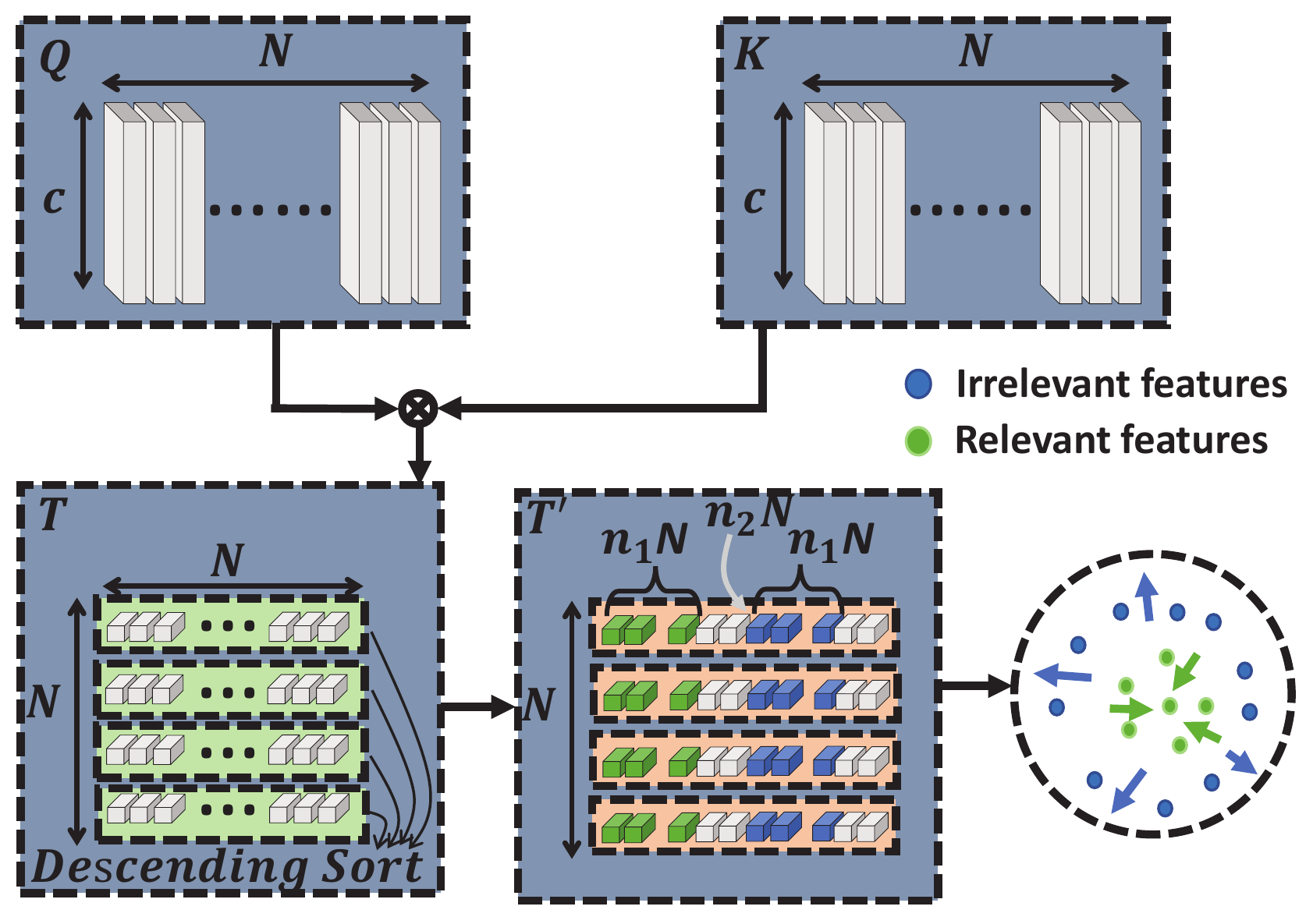} 
	\caption{The illustration of our contrastive learning scheme for ENLCA. For each ordered sequence, we take the top $n_{1}N$ related ones as relevant features and $n_{1}N$ unrelated ones starting from $n_{2}N$ as irrelevant features.  }
	\label{fig:contrastive_method}
\end{figure}

To alleviate the problem, we further develop Efficient Non-Local Contrastive Attention (ENLCA) by applying contrastive learning. The goal of adopting contrastive learning is to increase the gap between irrelevant and relevant features. As shown in Figure \ref{fig:contrastive_method}, Contrastive Learning loss ${\cal L}_{cl}$ for training ENLCA can be formulated as:
\begin{equation} 
\boldsymbol{T}_{i,j} = k\frac{\boldsymbol{Q}_{i}^{\top}}{\|\boldsymbol{Q}_{i}\|} \frac{\boldsymbol{K}_{j}}{\|\boldsymbol{K}_{j}\|}, k>1, \boldsymbol{T}_{i,j} \in \boldsymbol{T},
\end{equation}
\begin{equation} 
\boldsymbol{T}_{i}^{\prime} = \operatorname{sort}\left(\boldsymbol{T}_{i},\operatorname{Descending}\right) , \boldsymbol{T}_{i}^{\prime} \in \boldsymbol{T}^{\prime},
\boldsymbol{T}_{i} \in \boldsymbol{T},
\end{equation}
\begin{equation}
{\cal L}_{cl}=\frac{1}{N}\sum_{i=1}^{N}-\log \frac{\sum_{j=1}^{n_{1}N}\exp \left( \boldsymbol{T}_{i,j}^{\prime}   \right)/ n_{1}N}{\sum_{j=n_{2}N}^{(n_{1}+n_{2})N} \exp \left(\boldsymbol{T}_{i,j}^{\prime} \right)/ n_{1}N }+b,
\end{equation}
where $N$ indicates the input size. $b$ is a margin constant. $n_{1}$ represents the percentage of relevant and irrelevant features in the feature map, and $n_{2}$ is the start index percentage for irrelevant features in the feature map, respectively. $\boldsymbol{T}_{i,j}$ measures relevance between $\boldsymbol{Q}_{i}$ and $\boldsymbol{K}_{j}$ by normalized inner product. $\boldsymbol{T}_{i}$ and $\boldsymbol{T}_{i}^{\prime}$ stand for $i$-th row of $\boldsymbol{T}$ and  $\boldsymbol{T}^{\prime} \in \mathbb{R}^{N \times N}$ separately. Besides, $\boldsymbol{T}_{i}^{\prime}$ is descending sort result of $\boldsymbol{T}_{i}$.

Consequently, the overall loss function of our model is ultimately designed as:
\begin{equation}
{\cal L}_{rec}=\left\|I^{HR}-I^{SR}\right\|_{1},
\end{equation}
\begin{equation}
{\cal L} = {\cal L}_{rec}+\lambda_{cl} {\cal L}_{cl},
\end{equation}
where ${\cal L}_{rec}$ is Mean Absolute Error (MAE) aiming to reduce distortion between the predicted SR image $I_{SR}$ and the target HR image $I_{HR}$, and the weight $\lambda_{cl}$ is 1e-3.

\subsection{Efficient Non-Local Contrastive Network}

To demonstrate the effectiveness of our ENLCA module, we integrate it into the EDSR, a simple SR network consisting of 32 residual blocks, to form the Efficient Non-Local Contrastive Network (ENLCN). As shown in Figure \ref{fig:all_process},  ENLCN uses five ENLCA modules with one insertion after every eight residual blocks.

\begin{table*}[htb]
	\centering
	\caption{Quantitative results on benchmark datasets. Best and second best results are colored with  \textcolor{red}{red}  and  \textcolor{blue}{blue}.}
	\begin{tabular}{|c|c|c|c|c|c|c|}
		\hline
		\multirow{2}[4]{*}{Method} & \multirow{2}[4]{*}{Scale} & Set5 & Set14 & B100 & Urban100 & Manga109\\
		\cline{3-7}     &  & PSNR/SSIM  & PSNR/SSIM & PSNR/SSIM & PSNR/SSIM & PSNR/SSIM\\
		\hline
		\hline
		LapSRN  & ×2    & 37.52  / 0.9591  & 33.08  / 0.9130  & 31.08  / 0.8950  & 30.41  / 0.9101  & 37.27  / 0.9740  \\
		MemNet  & ×2    & 37.78  / 0.9597  & 33.28  / 0.9142  & 32.08  / 0.8978  & 31.31  / 0.9195  & 37.72  / 0.9740  \\
		SRMDNF  & ×2    & 37.79 / 0.9601  & 33.32  / 0.9159  & 32.05  / 0.8985  & 31.33  / 0.9204  & 38.07  / 0.9761  \\
		DBPN  & ×2    & 38.09  / 0.9600  & 33.85  / 0.9190  & 32.27  / 0.9000  & 32.55  / 0.9324  & 38.89  / 0.9775  \\
		RDN   & ×2    & 38.24  / 0.9614  & 34.01  / 0.9212  & 32.34  / 0.9017  & 32.89  / 0.9353  & 39.18  / 0.9780  \\
		RCAN  & ×2    & 38.27  / 0.9614  & \textcolor{blue}{34.12}  / 0.9216  & 32.41  / 0.9027  & 33.34  / 0.9384  & 39.44  / 0.9786  \\
		NLRN  & ×2    & 38.00  / 0.9603  & 33.46  / 0.9159  & 32.19  / 0.8992  & 31.81  / 0.9249  & –  \\
		RNAN  & ×2    & 38.17  / 0.9611  & 33.87  / 0.9207  & 32.32  / 0.9014  & 32.73  / 0.9340  & 39.23  / 0.9785  \\
		SRFBN  & ×2    & 38.11  / 0.9609  & 33.82  / 0.9196  & 32.29  / 0.9010  & 32.62  / 0.9328  & 39.08  / 0.9779  \\
		OISR  & ×2    & 38.21  / 0.9612  & 33.94  / 0.9206  & 32.36  / 0.9019  & 33.03  / 0.9365  & –  \\
		SAN   & ×2    & 38.31  / \textcolor{red}{0.9620}  & 34.07  / 0.9213  & 32.42  / \textcolor{blue}{0.9028}  & 33.10  / 0.9370  & 39.32  / \textcolor{red}{0.9792}  \\
		
		NLSN  & ×2    & \textcolor{blue}{38.34}  / 0.9617  & 34.08  / \textcolor{red}{0.9231}  & \textcolor{blue}{32.43}  / 0.9027  & \textcolor{blue}{33.42}  / \textcolor{blue}{0.9394}  & \textcolor{blue}{39.59}  / 0.9789  \\
		\hline
		EDSR  & ×2    & 38.11  / 0.9602  & 33.92  / 0.9195  & 32.32  / 0.9013  & 32.93  / 0.9351  & 39.10  / 0.9773  \\
		ENLCN (ours) & ×2    & \textcolor{red}{38.37}  / \textcolor{blue}{0.9618}  & \textcolor{red}{34.17}  / \textcolor{blue}{0.9229}  & \textcolor{red}{32.49}  / \textcolor{red}{0.9032}  & \textcolor{red}{33.56}  / \textcolor{red}{0.9398} & \textcolor{red}{39.64}  / \textcolor{blue}{0.9791}  \\
		\hline
		\hline
		LapSRN  & ×4    & 31.54  / 0.8850  & 28.19 / 0.7720  & 27.32  / 0.7270  & 25.21  / 0.7560  & 29.09  / 0.8900  \\
		MemNet  & ×4    & 31.74  / 0.8893  & 28.26  / 0.7723  & 27.40  / 0.7281  & 25.50  / 0.7630  & 29.42  / 0.8942  \\
		SRMDNF  & ×4    & 31.96  / 0.8925  & 28.35  / 0.7787  & 27.49  / 0.7337  & 25.68  / 0.7731  & 30.09  / 0.9024  \\
		DBPN  & ×4    & 32.47  / 0.8980  & 28.82  / 0.7860  & 27.72  / 0.7400  & 26.38  / 0.7946  & 30.91  / 0.9137  \\
		RDN   & ×4    & 32.47  / 0.8990  & 28.81  / 0.7871  & 27.72  / 0.7419  & 26.61  / 0.8028  & 31.00 / 0.9151  \\
		RCAN  & ×4    & 32.63  / 0.9002  & 28.87  / 0.7889  & 27.77  / 0.7436  & 26.82  / 0.8087  & 31.22  / 0.9173  \\
		NLRN  & ×4    & 31.92  / 0.8916  & 28.36  / 0.7745  & 27.48  / 0.7306  & 25.79  / 0.7729  &   - \\
		RNAN  & ×4    & 32.49  / 0.8982 & 28.83  / 0.7878  & 27.72  / 0.7421  & 26.61  / 0.8023  & 31.09  / 0.9149  \\
		SRFBN  & ×4    & 32.47  / 0.8983  & 28.81  / 0.7868  & 27.72  / 0.7409  & 26.60  / 0.8015  & 31.15  / 0.9160  \\
		OISR  & ×4    & 32.53  / 0.8992  & 28.86  / 0.7878  & 27.75  / 0.7428  & 26.79  / 0.8068  &   - \\
		SAN   & ×4    & \textcolor{blue}{32.64}  / \textcolor{blue}{0.9003}  & \textcolor{blue}{28.92}  / 0.7888  & \textcolor{blue}{27.78}  / 0.7436  & 26.79  / 0.8068  & 31.18  / 0.9169  \\
		NLSN  & ×4    & 32.59  / 0.9000  & 28.87  / \textcolor{blue}{0.7891}  & \textcolor{blue}{27.78}  / \textcolor{blue}{0.7444}  & \textcolor{blue}{26.96}  / \textcolor{blue}{0.8109}  & \textcolor{blue}{31.27}  / \textcolor{blue}{0.9184}  \\
		\hline
		EDSR  & ×4    & 32.46  / 0.8968  & 28.80  / 0.7876  & 27.71  / 0.7420  & 26.64  / 0.8033  & 31.02  / 0.9148  \\
		ENLCN (ours) & ×4    & \textcolor{red}{32.67}  / \textcolor{red}{0.9004}  & \textcolor{red}{28.94}  / \textcolor{red}{0.7892}  & \textcolor{red}{27.82}  / \textcolor{red}{0.7452}  & \textcolor{red}{27.12}  / \textcolor{red}{0.8141}  & \textcolor{red}{31.33}  / \textcolor{red}{0.9188}  \\
		\hline
	\end{tabular}%
	\label{tab:quantitative}%
\end{table*}%

\begin{figure*}[htb]
	\centering
	\includegraphics[height=9.56cm,width=18cm]{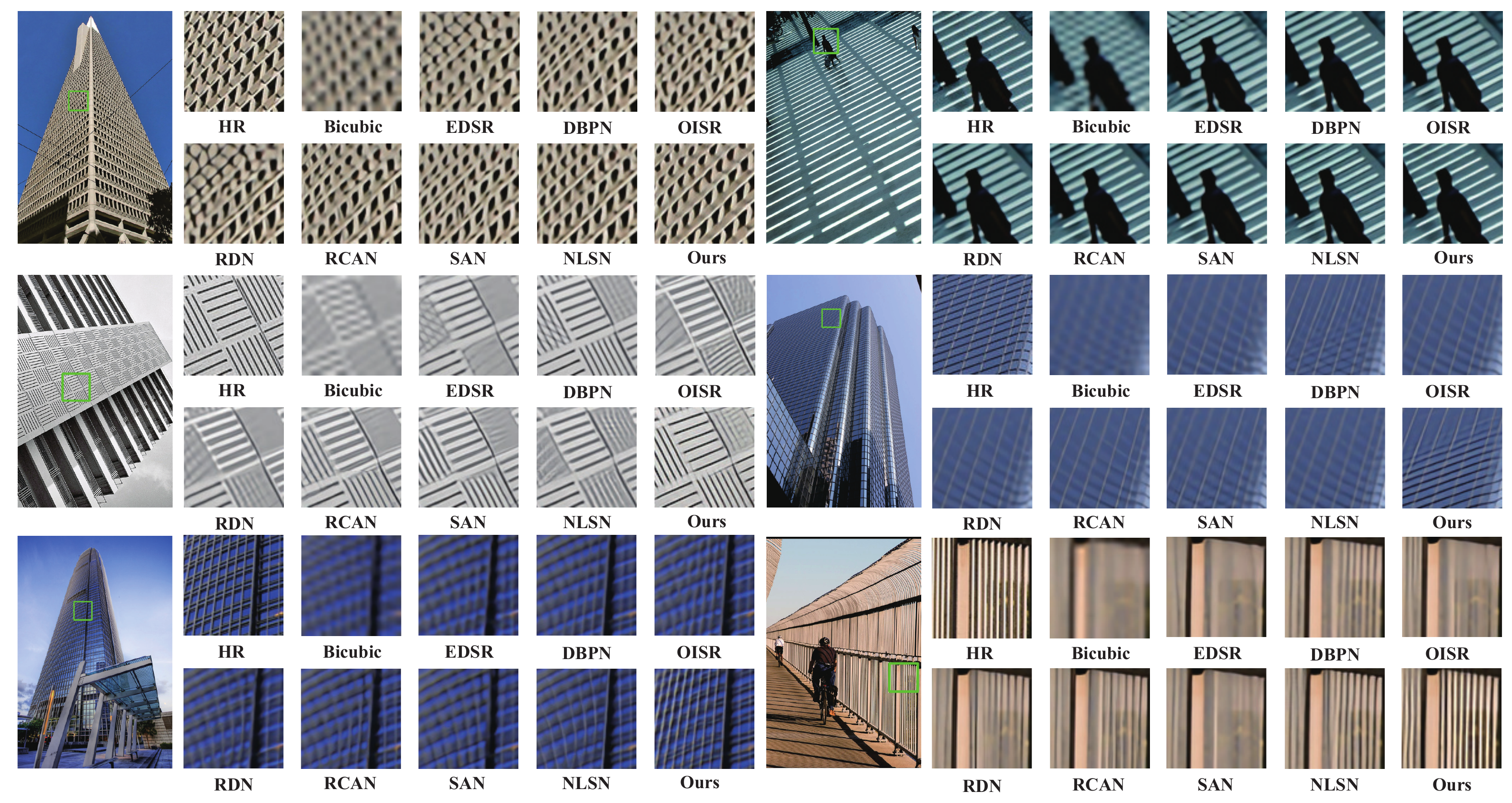} 
	\caption{Visual comparison for $4\times$ SR on Urban100 dataset. For all the shown examples,  our method significantly outperforms other state-of-the-arts, particularly in the image rich in repeated textures and structures.}
	\label{fig:exp}
\end{figure*}
\section{Experiments}

\subsection{Datasets and Evaluation Metrics}
Following EDSR \cite{EDSR} and RNAN \cite{RNAN}, we use DIV2K \cite{DIV2K}, a dataset consists of 800 training images, to train our models. We test our
method on 5 standard SISR benchmarks: Set5 \cite{Set5}, Set14 \cite{Set14}, B100 \cite{B100}, Urban100 \cite{Urban100} and Manga109 \cite{Manga109}. We evaluate all the SR results by PSNR and SSIM metrics on Y channel only in the transformed YCbCr space.

\subsection{Implementation details}
For ENLCA, we regenerate Gaussian random matrix $\boldsymbol{F}$ every epoch. Additionally, amplification factor $k$ is 6, and margin $b$ is 1. The number of random samples $m$ is set to 128. We build ENLCN using EDSR backbone with 32-residual blocks and 5 additional ENLCA blocks. All convolutional kernel size in the network is $3\times3$. All intermediate features have 256 channels except for those embedded features in the attention module having 64 channels. The last convolution layer has 3 filters to transform the feature map into a 3-channel RGB image.

During training, we set  $n_{1}$ and $n_{2}$ for contrastive learning to 2$\%$ and 8$\%$, separately. Besides, we randomly crop $28 \times 28$ and $46 \times 46$ patches from 16 images to form an input batch for $\times$4 and $\times$2 SR, respectively. We augment the training patches by randomly horizontal flipping and rotating $90^{\circ}$, $180^{\circ}$, $270^{\circ}$. The model is optimized by ADAM optimizer \cite{adam}
with $\beta_{1}= 0.9$, $\beta_{2}=0.99$ and initial learning rate of 1e-4. We reduce the learning rate by 0.5 after 200 epochs and obtain the final model after 1000 epochs. We first warm up the network via training 150 epochs with only ${\cal L}_{rec}$, then train with all loss functions. The model is implemented with PyTorch and trained on Nvidia 2080ti GPUs.

\subsection{Comparisons with State-of-the-arts}

To validate the effectiveness of our ENLCA, we compare our approach with 13 state-of-the-art methods,
which are LapSRN \cite{LapSRN}, SRMDNF \cite{SRMDNF}, MemNet \cite{Memnet}
, EDSR \cite{EDSR}, DBPN \cite{DBPN}, RDN \cite{RDN}, RCAN \cite{RCAN},
NLRN \cite{NLRN}, SRFBN \cite{SRFBN}, OISR \cite{OISR}, SAN \cite{SAN} and NLSN \cite{NLSN}.

In Table \ref{tab:quantitative}, we display the quantitative comparisons of scale factor $\times2$ and $\times4$. Compared with other methods, our ENLCN achieves the best results on almost all benchmarks and all scale factors. It is worth noting that adding additional ENLCAs brings significant improvement and even drives backbone EDSR outperforming the state-of-the-art methods, such as SAN and RCAN. Specifically, compared with EDSR, ENLCN improves about 0.2 dB in Set5, Set14, and B100 while around 0.5 dB in Urban100 and Manga109. Furthermore, compared with previous non-local based methods such as NLRN and RNAN, our network shows a huge superiority in performance. This is mainly because ENLCA only focuses on relevant features aggregation and filters out the noisy information from irrelevant features, which yields a more accurate prediction. Moreover, compared with the Sparse Non-Local Attention (NLSA) based method like NLSN, our ENLCN embodies advance in almost all entries. This is because NLSA aggregates relevant information roughly and may ignore important information, while ENLCA aggregates all relevant information. It is noted that all these improvements merely cost a small amount of computation, which roughly equals the computation of a convolution operation. The qualitative evaluations on Urban100 are shown in Figure \ref{fig:exp}.

\section{Ablation Study}
In this section, we conduct experiments to investigate our ENLCA. We build the baseline model with 32 residual blocks and insert corresponding attention variants after every 8 residual blocks. 
\begin{table}
	\centering
	\caption{Ablation experiments conducted on Set5 ($\times$4) to
		study the effectiveness of the proposed Efficient Non-Local Attention (ENLA) module, multiplying amplification factor $k$, and contrastive learning.  }
	\begin{tabular}{|cccc|c|}
		\hline
		Base & ENLA & $k$  & contrastive learning & PSNR \\
		\hline
		\hline
		\Checkmark     & \XSolidBrush      &   \XSolidBrush    &   \XSolidBrush    & 32.21 \\
		\hline
		\Checkmark     & \Checkmark     &   \XSolidBrush    &   \XSolidBrush    & 32.37 \\
		\hline
		\Checkmark     & \Checkmark     & \Checkmark     &   \XSolidBrush    & 32.44 \\
		\hline
		\Checkmark     & \Checkmark     &   \XSolidBrush    & \Checkmark     & 32.41 \\
		\hline
		\Checkmark     & \Checkmark     & \Checkmark     & \Checkmark     & 32.48 \\
		\hline
	\end{tabular}%
	\label{tab:ablation}%
\end{table}%

\textbf{Efficient Non-Local Contrastive Attention module.} To demonstrate the effectiveness of the proposed Efficient Non-Local Contrastive Attention (ENLCA) module, we construct a baseline model by progressively adding our attention module or sparsity schemes. As shown in Table \ref{tab:ablation}, comparing the first and the second row, Our ENLA brings 0.16 dB improvement over baseline, which demonstrates the effectiveness of ENLA. Furthermore, by solely adding the Sparsity Aggregation scheme such as multiplying the input by an amplification factor $k$ in Eq \ref{Eq:sparse} and contrastive learning, results further improve around 0.07 dB. In the last row, we combine all modules and Sparsity Aggregation schemes, achieving an 0.27 dB improvement over the baseline.

These facts demonstrate that single Efficient Non-Local Attention (ENLA) works well in SISR. Besides, adding $k$ and contrastive learning to focus attention on the most informative positions is essential.

\begin{table}
	\centering
	\caption{The experiment conducted on Set5 ($\times$4) to explore the effects of $n_{1} (\%) $ and $n_{2} (\%) $ on contrastive learning.}
	\resizebox{0.8\linewidth}{!}{
	\begin{tabular}{|c||c|c|c|c|}
		\hline
		\diagbox{$n_{1}$}{PSNR}{$n_{2}$} & 4   & 8 & 13   & 20 \\
		\hline
		\hline
		1     & 32.40 & 32.42 & 32.41 & 32.38 \\
		\hline
		1.5    & 32.42 & 32.47  & 32.44  & 32.42 \\
	\hline
		2    & 32.45  & 32.48 & 32.44 & 32.42 \\
	\hline
		3    & 32.43 & 32.46 & 32.44 & 32.41 \\
		\hline
	\end{tabular}%
}
	\label{tab:n1 and n2}%
\end{table}%

\textbf{The effect of $n_{1}$ and $n_{2}$ on contrastive learning.} The $n_{1}$ determines the percentage of relevant and irrelevant features participating in contrastive learning, and $n_{2}$ indicates the start index percentage of irrelevant features in the feature map. Results of the models trained on DIV2K ($\times$4, 200 epochs) and evaluated on Set5 ($\times$4) with different $n_{1}$ and $n_{2}$ are presented in Table \ref{tab:n1 and n2}. When $n_{1}=2\%$ and $n_{2}=8\%$, the model achieves the best performance. In the second column, with $n_{1}$ increase from 1$\%$ to 2$\%$, the PSNR improves for taking more informative features as relevant features and uninformative features as irrelevant features. However, as $n_{1}$ increases from 2$\%$ to 3$\%$, the performance decreases for fusing noisy features as relevant features.  Similarly, in second rows, $n_{2}$ increases from 4$\%$ to 8$\%$ bringing improvements on the metric for taking less relative informative features as irrelevant features and increases from 8$\%$ to 20$\%$ leading to degradation of performance for missing taking noisy features as irrelevant features.

\begin{figure}
	\centering
	\includegraphics[height=6cm,width=8cm]{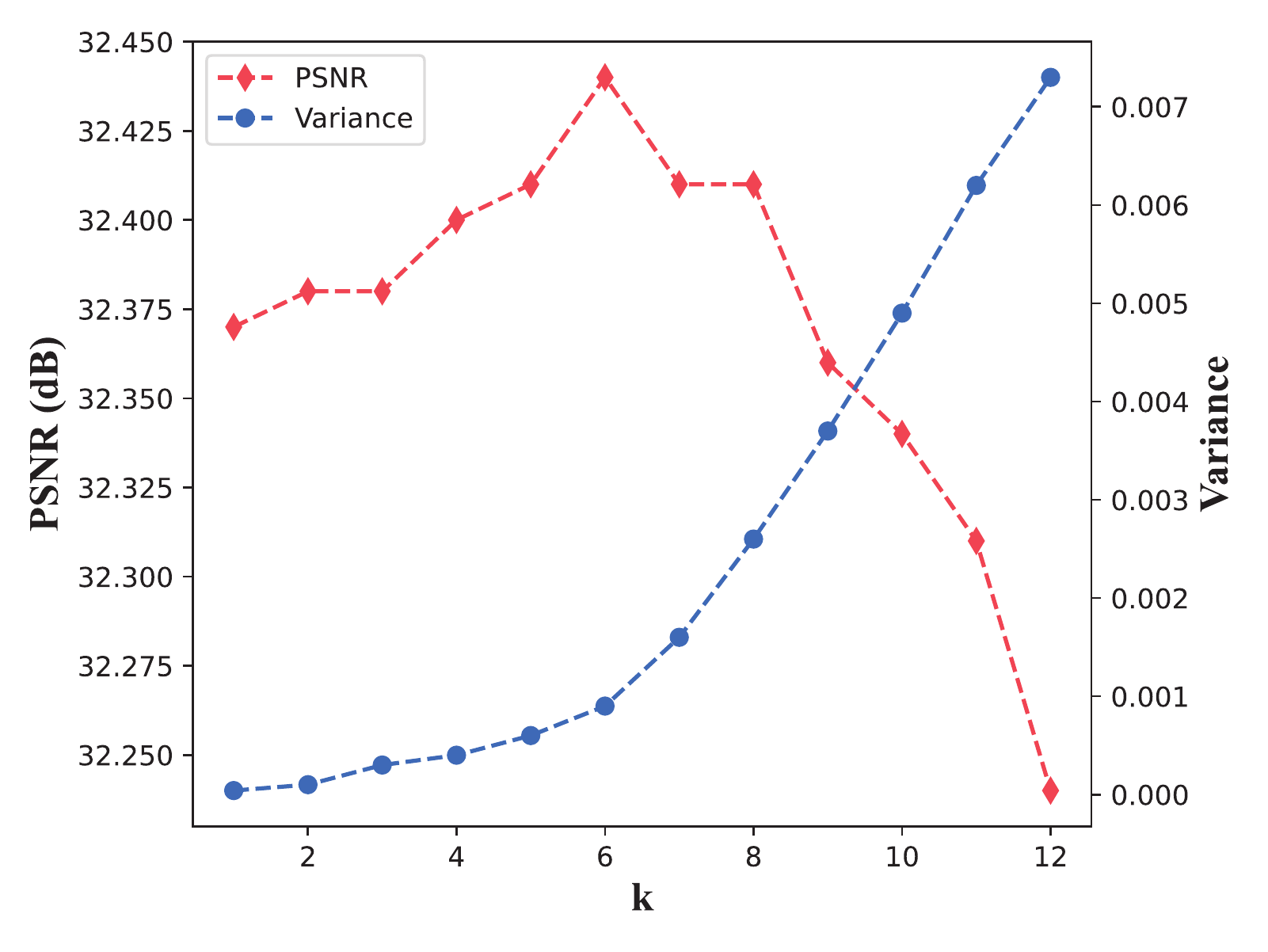} 
	\caption{The relationship on amplification factor $k$ with SR results and approximation variance of ENLA. }
	\label{fig:mse}
\end{figure}

\textbf{The relationship on $k$ with SR results and approximation variance.} We conduct the experiment on Set5 ($\times$4). As shown in Figure \ref{fig:mse}, when $k$ is set to 6, the model achieves the best performance. That is mainly because increasing $k$ causes the amplification of $\boldsymbol{Q}$ and $\boldsymbol{K}$, giving irrelevant features lower weight and relevant features larger weight in feature aggregation, but also results in the approximation variance of ENLA increasing exponentially. Hence, it is of crucial importance to find the right $k$ to balance the merits and demerits.  

\begin{figure}
	\centering
	\includegraphics[height=2.4cm,width=8cm]{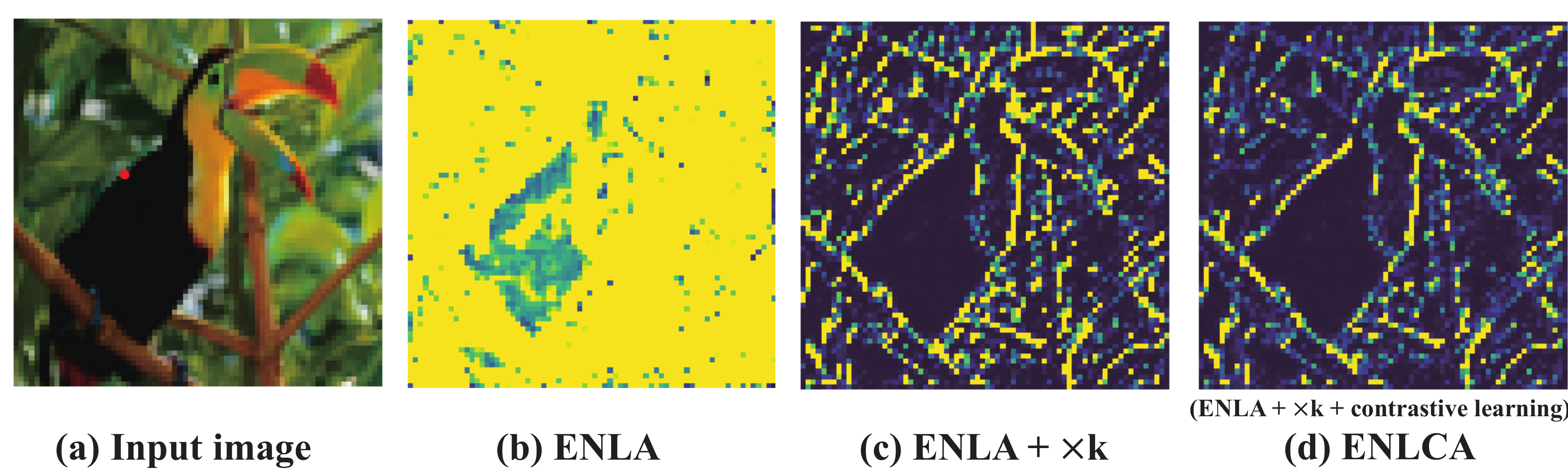} 
	\caption{The verification of the sparsity bringing by Sparse Aggregation.  }
	\label{fig:A_heatmap}
\end{figure}  

\textbf{The effectiveness of Sparse Aggregation.} To verify the sparsity bringing by Sparse Aggregation, we progressively add amplification factor $k$ and contrastive learning on ENLA and visualize the corresponding correlation maps obtained by softmax and inner product between the marked red feature and all features. As shown in Figure \ref{fig:A_heatmap}, ENLA gives large weights to unrelated regions. By multiplying the input by $k$, ENLA strengthens the effect of relevant features and suppresses irrelevant contents. When ENLA further adopts contrastive learning, the weights of relevant and irrelevant features are further distinct.  

\begin{table}
	\centering
	\caption{Efficiency and performance comparison on Urban100 ($\times$4). $m$
		is the number of random samples.}
	\resizebox{0.6\linewidth}{!}{
	\begin{tabular}{|c|c|c|c|c|c|}
		\hline	
		Methods & GFLOPs & PSNR   \\
		\hline
		\hline
		Baseline & 0     & 26.64 \\ 
		NLA   & 25.60  & 26.87 \\ 
		Conv  & 0.74  & -    \\ 
		NLSA-r4 & 5.08  & 26.96 \\ 
		\hline
		ENLCA-$m$256 & 1.31  & 27.12 \\ 
		ENLCA-$m$128 & 0.66  & 27.12 \\ 
		ENLCA-$m$64 & 0.33  & 27.09 \\
		ENLCA-$m$32 & 0.16  & 27.07 \\
		ENLCA-$m$16 & 0.08  & 27.07 \\
		ENLCA-$m$8 & 0.04  & 27.03 \\
		ENLCA-$m$4 & 0.02  & 27.01 \\
		ENLCA-$m$2 & 0.01  & 26.88 \\
		\hline
	\end{tabular} 
	\label{tab:efficiency}%
}
\end{table}%

\textbf{Efficiency.} We compare the proposed ENLCA with standard Non-Local Attention (NLA) in terms of computational efficiency. Table
\ref{tab:efficiency} shows the computational cost and the corresponding performances. The input is assumed to be the size
of 100 $\times$ 100, and both input and output channels are 
64. We also add a normal 3×3 convolution operation
for better illustration. As shown in Table \ref{tab:efficiency}, ENLCA significantly reduces the computational cost of the NLA. Even compared with NLSA, our ENLCA achieves superior performance with much less computational cost. For example, our ENLCA-$m$16 brings 0.11 dB improvement on performance and is 100 times more efficient than NLSA. Furthermore, it is notable that our ENLCA has negligible computational cost compared with a convolution but achieves better performances than NLA and NLSA \cite{NLSN}. The best result is achieved by ENLCA-$m$128 and ENLCA-$m$256, which shows a bottleneck for increasing $m$ to improve performance.

\section{Conclusion}
In this paper, we propose a novel Efficient Non-Local Contrastive Attention (ENLCA) to enable effective and efficient long-range modeling for deep single image super-resolution networks. To reduce the excessive computational cost of non-local attention, we propose an Efficient Non-Local Attention (ENLA)  by exploiting the kernel method to approximate exponential function. Furthermore, ENLCA adopts Sparse Aggregation, including multiplying inputs by an amplification factor and adding contrastive learning to focus on the most related locations and ignore unrelated regions. Extensive experiments on several benchmarks demonstrate the superiority of ENLCA, and comprehensive ablation analyses verify the effectiveness of ENLA and Sparse Aggregation.

\section{Acknowledgments}
This work was partly supported by the Natural Science Foundation of China (No.62171251), the Natural Science Foundation of Guangdong Province (No.2020A1515010711), the Special Foundation for the Development of Strategic Emerging Industries of Shenzhen (No.JCYJ20200109143010272) and Oversea Cooperation Foundation of Tsinghua Shenzhen International Graduate School.

\begin{appendix}
	\maketitle

	\appendix
	\section{Appendix}
	
	\subsection{Proof 1}
	\label{subsection:A}
	In this section, we provide proof that our Efficient Non-Local Contrastive Attention (ENLCA) is the unbiased approximation of standard non-local attention.

	First, we define $\boldsymbol{f} \in \mathbb{R}^{c}$ and use the fact that for any $\boldsymbol{Q}_{i}, \boldsymbol{K}_{j}, \in \mathbb{R}^{c}$,
	\begin{equation}
	(2 \pi)^{-c / 2} \int \exp \left(-\|\boldsymbol{f}-\left(\boldsymbol{Q}_{i}+\boldsymbol{K}_{j}\right) \|^{2} / 2 \right) d \boldsymbol{\boldsymbol{f}}=1.
	\label{Eq:1}
	\end{equation}

	Then, by using Eq.\ref{Eq:1}, we can derive the approximation of $\exp \left(\|\boldsymbol{Q}_{i}+\boldsymbol{K}_{j}\|^{2}/2\right)$ as follows :
	\begin{equation}
	\begin{aligned}
	&\exp \left(\|\boldsymbol{Q}_{i}+\boldsymbol{K}_{j}\|^{2}/2\right)=(2 \pi)^{-c / 2} \exp  \left(\|\boldsymbol{Q}_{i}+\boldsymbol{K}_{j}\|^{2} / 2\right) \\
	&\cdot \int \exp \left(-\|\boldsymbol{f}-(\boldsymbol{Q}_{i}+\boldsymbol{K}_{j})\|^{2} / 2\right) d \boldsymbol{f} \\
	&=(2 \pi)^{-c / 2} \int \exp (-\|\boldsymbol{f}\|^{2} / 2+\boldsymbol{f}^{\top}(\boldsymbol{Q}_{i}+\boldsymbol{K}_{j})\\
	&-\|\boldsymbol{Q}_{i}+\boldsymbol{K}_{j}\|^{2} / 2+\|\boldsymbol{Q}_{i}+\boldsymbol{K}_{j}\|^{2} / 2) d \boldsymbol{f} \\
	&=(2 \pi)^{-c / 2} \int \exp \left(-\|\boldsymbol{f}\|^{2} / 2+\boldsymbol{f}^{\top}(\boldsymbol{Q}_{i}+\boldsymbol{K}_{j})\right) d \boldsymbol{f} \\
	&=(2 \pi)^{-c / 2} \int \exp \left(-\|\boldsymbol{f}\|^{2} / 2\right) \cdot \exp \left(\boldsymbol{f}^{\top}(\boldsymbol{Q}_{i}+\boldsymbol{K}_{j})\right) d \boldsymbol{f} \\
	&=\mathbb{E}_{\boldsymbol{f} \sim \mathcal{N}\left(\boldsymbol{0}_{c}, \boldsymbol{I}_{c}\right)} \left[\exp \left(\boldsymbol{f}^{\top}\left( \boldsymbol{Q}_{i}+\boldsymbol{K}_{j}\right)\right)\right].
	\end{aligned}
	\label{Eq:2}
	\end{equation}
	
	By exploiting the approximation form in Eq.\ref{Eq:2} to obtain the unbiased approximation of $\exp \left(\boldsymbol{Q}_{i}^{\top} \boldsymbol{K}_{j}\right)$ as
	\begin{equation}
	\begin{aligned}
	&\mathrm{K}(\boldsymbol{Q}_{i}, \boldsymbol{K}_{j})=\exp \left(\boldsymbol{Q}_{i}^{\top} \boldsymbol{K}_{j}\right)\\
	&= \exp \left(-\left(\|\boldsymbol{Q}_{i}\|^{2} + \| \boldsymbol{K}_{j} \|^{2}\right)/2\right) \cdot \exp \left( \|\boldsymbol{Q}_{i}+\boldsymbol{K}_{j}\|^{2}/2\right)\\
	& = \mathbb{E}_{\boldsymbol{f} \sim \mathcal{N}\left(\boldsymbol{0}_{c}, \boldsymbol{I}_{c}\right)} \exp \left(\boldsymbol{f}^{\top}\left( \boldsymbol{Q}_{i}+\boldsymbol{K}_{j}\right)-\frac{\|\boldsymbol{Q}_{i}\|^{2}+\|\boldsymbol{K}_{j}\|^{2}}{2}\right) \\
	&= \phi(\boldsymbol{Q}_{i})^{T}\phi(\boldsymbol{K}_{j}).
	\end{aligned}
	\label{Eq:3}
	\end{equation}

	\subsection{Proof 2}
	\label{subsection:B}
	In this section, we provide the proof of approximation variance of ENLCA. Here, we denote: $\boldsymbol{z} = \boldsymbol{Q}_{i}+\boldsymbol{K}_{j}$.
	
	From Eq.\ref{Eq:3}, we can know that:
	\begin{equation}
	\mathbb{E}_{\boldsymbol{f} \sim \mathcal{N}\left(\boldsymbol{0}_{c}, \mathbf{I}_{c}\right)}\left[\exp \left(\boldsymbol{f}^{\top} \boldsymbol{z}\right)\right]=\exp \left(\frac{\|\boldsymbol{z}\|^{2}}{2}\right).
	\end{equation}
	
	From Eq.\ref{Eq:3}, we can further derive the variance of approximation of $\exp \left(\boldsymbol{Q}_{i}^{\top} \boldsymbol{K}_{j}\right)$ as:   
	\begin{equation}
	\begin{aligned}
	&\operatorname{Var}\left(\phi(\boldsymbol{Q}_{i})^{T}\phi(\boldsymbol{K}_{j})\right)\\
	&=\frac{1}{m} \exp \left(-\left(\|\boldsymbol{Q}_{i}\|^{2}+\|\boldsymbol{K}_{j}\|^{2}\right)\right) \cdot \operatorname{Var}\left(\exp \left(\boldsymbol{f}^{\top} \boldsymbol{z}\right)\right) \\
	&=\frac{1}{m} \exp \left(-\left(\|\boldsymbol{Q}_{i}\|^{2}+\|\boldsymbol{K}_{j}\|^{2}\right)\right)\\
	&\cdot \left(\mathbb{E}\left[\exp \left(2 \boldsymbol{f}^{\top} \boldsymbol{z}\right)\right]-\left(\mathbb{E}\left[\exp \left(\boldsymbol{f}^{\top} \boldsymbol{z}\right)\right]\right)^{2}\right) \\
	&=\frac{1}{m} \exp \left(-\left(\|\boldsymbol{Q}_{i}\|^{2}+\|\boldsymbol{K}_{j}\|^{2}\right)\right) \\ 
	&\cdot \left(\exp \left(2\|\boldsymbol{z}\|^{2}\right)-\exp \left(\|\boldsymbol{z}\|^{2}\right)\right)\\
	&=\frac{1}{m} \mathrm{K}^{2}(\boldsymbol{Q}_{i}, \boldsymbol{K}_{j})\cdot\left(\exp \left(\|\boldsymbol{z}\|^{2}\right)-1\right)\\
	&=\frac{1}{m} \mathrm{K}^{2}(\boldsymbol{Q}_{i}, \boldsymbol{K}_{j})\cdot\left(\exp \left(\|\boldsymbol{Q}_{i}+\boldsymbol{K}_{j}\|^{2}\right)-1\right).
	\end{aligned}
	\end{equation}
	\begin{figure*}[h]
		\centering
		\includegraphics[height=8.6cm,width=16cm]{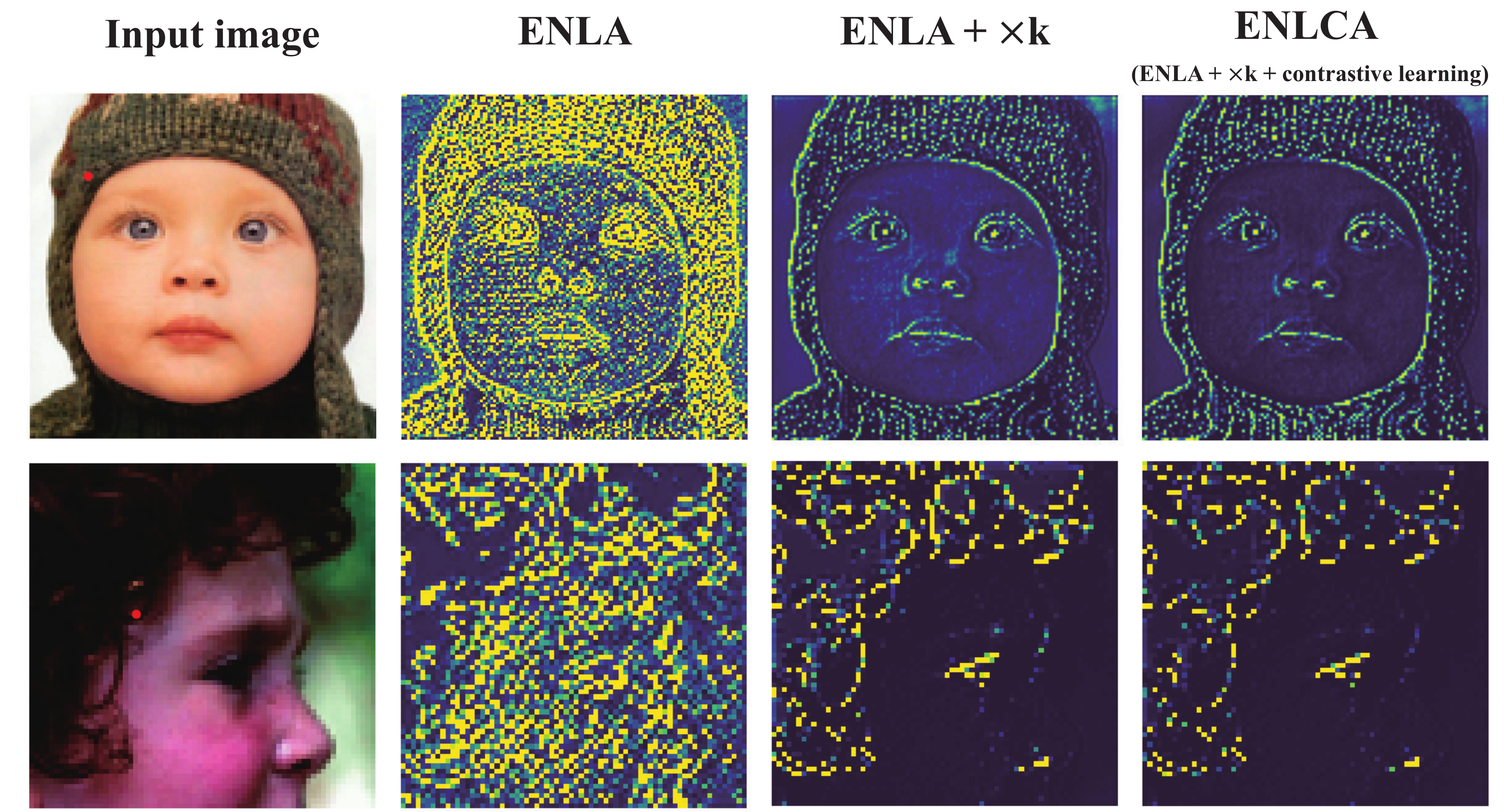} 
		\caption{More visualization of the sparsity bringing by Sparse
			Aggregation. On the top of the ENLA module, multiplying the amplification factor $k$ ($\times k$) and contrastive learning are progressively added. The correlation maps are obtained by softmax and inner product between the marked red feature and all features.}
		\label{fig:more_ablation}
	\end{figure*}
	
	\subsection{Running Time Comparison}
	\label{subsection:C}
	Here we present the running time comparison with standard Non-Local Attention
	(NLA) and Non-Local Sparse Attention (NLSA) \cite{NLSN}, following the settings in Table 4 of the paper. Models are evaluated at two input sizes: 100$\times$100 and 150$\times$150. We calculate the running time with the average of 1K times on one Nvidia 2080ti GPU. As shown in Table \ref{computation_cost}, compared with NLSA-\textit{r}4 on the input with size of 150$\times$150, our ENLCA-\textit{m}128 achieves around 0.2 dB improvement and significantly saves running time, demonstrating that ENLCA is indeed an efficient operation. 
	\begin{table}[h]
		\centering
		\caption{Running time comparison with NLA and NLSA.}
		\resizebox{1\linewidth}{!}{
			\begin{tabular}{c|ccc}
				\hline
				& 100$\times$100 (ms) & 150$\times$150 (ms) & PSNR (dB) \\
				\hline
				\hline
				NLA   & 42.8  & 107.2 & 26.87 \\
				\hline
				NLSA-\textit{r}4 & 3.9   & 8.8   & 26.96 \\
				\hline
				ENLCA-\textit{m}128 & 0.9     & 1.6   & 27.12 \\
				\hline
			\end{tabular}%
		}
		\label{computation_cost}%
	\end{table}%

	\subsection{Comparison with Criss-Cross Attention}
	\label{subsection:D}
	Criss-Cross Attention is a variant of non-local
	attention widely used for high-level vision tasks. The Criss-Cross Attention aggregates the features on the criss-cross path of the query point to realize sparse aggregation. Thus, the aggregation features of Criss-Cross Attention merely depend on location. As shown in Table \ref{tab:Criss-Cross}, ENLCA, NLSA and standard Non-Local Attention outperform Criss-Cross Attention. Besides, our ENLCA achieves the best performance. Compared with using sparsity in the fixed location, it is demonstrated that enforcing sparsity based on content relevance can better exploit the global relevant information.  
	
	\begin{table}[h]
		\centering
		\caption{Comparison with Criss-Cross Attention on Urban100 ($\times 4$).}
		\resizebox{1\linewidth}{!}{
			\begin{tabular}{|c|c|c|c|c|c|}
				\hline
				& baseline & Non-Local & Criss-Cross & NLSA  & ENLCA \\
				\hline
				PSNR  & 26.64 & 26.87 & 26.85 & 26.96 & 27.12 \\
				\hline
			\end{tabular}%
		}
		\label{tab:Criss-Cross}%
	\end{table}%

	\subsection{More Visualization of Correlation Maps}
	\label{subsection:E}
	In this paper, the proposed ENLCA consists of three major components: Efficient Non-Local Attention (ENLA) and Sparse Aggregation. The Sparse Aggregation includes Multiplying the input by an amplification factor $k$ ($\times k$) and contrastive learning. On the top of the ENLA, we progressively add
	the $\times k$ and contrastive learning. In Figure \ref{fig:more_ablation}, we show more visualization of correlation maps to demonstrate the effectiveness of Sparse Aggregation.

\end{appendix}

 \bibliography{aaai22.bib}

\end{document}